\documentclass[10pt,twocolumn,letterpaper]{article}

\usepackage[pagenumbers]{cvpr} 
\usepackage{m3}

\definecolor{cvprblue}{rgb}{0.21,0.49,0.74}
\usepackage[pagebackref,breaklinks,colorlinks,allcolors=cvprblue]{hyperref}

\def\paperID{10374} 
\def\confName{CVPR}
\def\confYear{2026}

\title{M$^3$KG-RAG: Multi-hop Multimodal Knowledge Graph-enhanced \\ Retrieval-Augmented Generation}

\author{
Hyeongcheol Park\textsuperscript{\normalfont 1} \quad
Jiyoung Seo\textsuperscript{\normalfont 1}\quad
Jaewon Mun\textsuperscript{\normalfont 1}\quad
Hogun Park\textsuperscript{\normalfont 2}\\
Wonmin Byeon\textsuperscript{\normalfont 3}\quad
Sung June Kim\textsuperscript{\normalfont 1}\quad
Hyeonsoo Im\textsuperscript{\normalfont 4}\quad
JeungSub Lee\textsuperscript{\normalfont 4}\quad
Sangpil Kim\textsuperscript{\normalfont 1}\thanks{Corresponding author.}\vspace{0.3em}\\
\textsuperscript{\normalfont 1}Korea University \quad
\textsuperscript{\normalfont 2}Sungkyunkwan University \quad
\textsuperscript{\normalfont 3}NVIDIA Research\quad
\textsuperscript{\normalfont 4}Hanwha Systems
\vspace{-1.5em}
}

\begin{document}
\maketitle

\begin{abstract}
Retrieval-Augmented Generation (RAG) has recently been extended to multimodal settings, connecting multimodal large language models (MLLMs) with vast corpora of external knowledge such as multimodal knowledge graphs (MMKGs). 
Despite their recent success, multimodal RAG in the audio-visual domain remains challenging due to 1) limited modality coverage and multi-hop connectivity of existing MMKGs, and 2) retrieval based solely on similarity in a shared multimodal embedding space, which fails to filter out off-topic or redundant knowledge.
To address these limitations, we propose M$^3$KG-RAG, a Multi-hop Multimodal Knowledge Graph-enhanced RAG that retrieves query-aligned audio-visual knowledge from MMKGs, improving reasoning depth and answer faithfulness in MLLMs.
Specifically, we devise a lightweight multi-agent pipeline to construct multi-hop MMKG (M$^3$KG), which contains context-enriched triplets of multimodal entities, enabling modality-wise retrieval based on input queries.
Furthermore, we introduce GRASP (Grounded Retrieval And Selective Pruning), which ensures precise entity grounding to the query, evaluates answer-supporting relevance, and prunes redundant context to retain only knowledge essential for response generation. 
Extensive experiments across diverse multimodal benchmarks demonstrate that M$^3$KG-RAG significantly enhances MLLMs’ multimodal reasoning and grounding over existing approaches.
Project website: \url{https://kuai-lab.github.io/cvpr2026m3kgrag/}
\vspace{-1.em}
\end{abstract}
\vspace{-0.5em}
\section{Introduction}
\label{sec:intro}
\vspace{-0.5em}
The advancements in Retrieval-Augmented Generation (RAG) have substantially improved the factual accuracy and faithfulness of large language models (LLMs) by connecting them to vast external knowledge corpora~\cite{lewis2020retrieval, sharma2025retrieval}.
Recently, graph-based RAG methods~\cite{edge2024local,guo2024lightrag,zhang2025survey} have further pushed the progress by supporting structured reasoning and precise, query-relevant retrieval.
However, extending these schemes to multimodal settings—jointly handling audio, visual, and textual signals—is non-trivial as heterogeneous inputs raise complexity, motivating designs that explicitly account for multimodal structure.
Recent work~\cite{lee2024multimodal, zha2024m2conceptbase, ren2025videorag} addresses these challenges with multimodal knowledge graphs (MMKGs) that organize cross-modal knowledge as entities and relations, thereby delivering query-relevant evidence to multimodal large language models (MLLMs)~\cite{achiam2023gpt, yao2024minicpm}.

\begin{figure}[t]
    \centering
    \includegraphics[width=\linewidth]{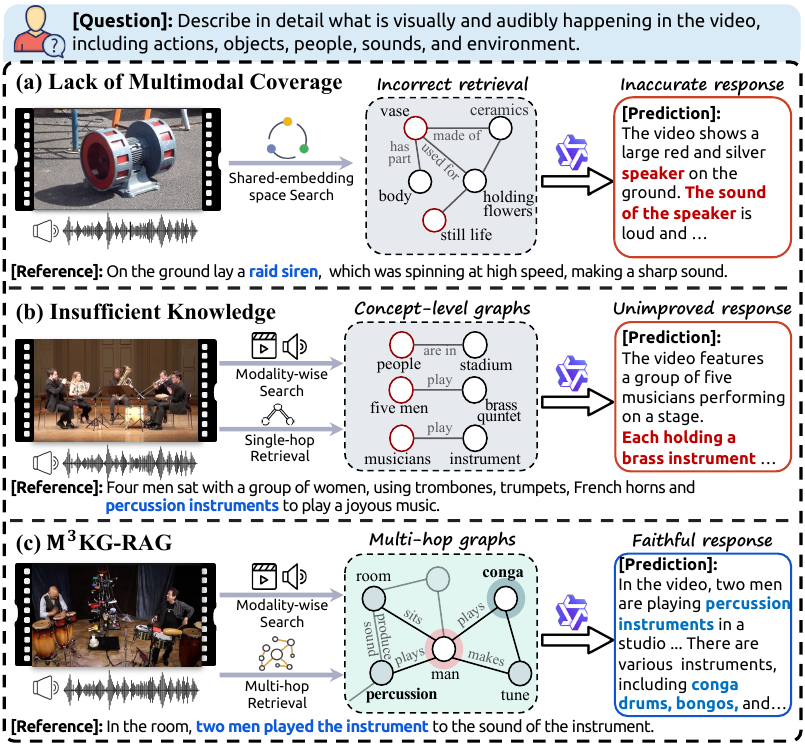}
    \caption{
    \textbf{Illustration of multimodal RAG scenarios.}
    Incorrect answers are shown in \textcolor{red}{red}, correct answers in \textcolor{blue}{blue}.
    (a) Shared embedding search misaligns with the audio-visual query.
    (b) Noisy, single-hop facts provide little answer support.
    (c) M$^3$KG-RAG uses modality-wise multi-hop retrieval for answer-supporting context.
    }
    \label{fig:intro}
    \vspace{-2.em}
\end{figure}

However, existing MMKG-enhanced RAG methods exhibit key limitations.
\textbf{First}, existing MMKGs~\cite{lee2023vista, zha2024m2conceptbase, liu2025aligning} largely emphasize image-text and provide limited audio-visual coverage, which hampers temporal and causal reasoning across modalities.
Moreover, the modality gap~\cite{ramasinghe2024accept} in unified multimodal embedding spaces makes cross-modal retrieval inaccurate~\cite{yeo2025universalrag}. 
As illustrated in \cref{fig:intro}-(a), matching an audio-visual query directly against a text-only knowledge base often fails to retrieve truly related evidence, which motivates modality-wise retrieval.
While recent work~\cite{park2025vat} builds an audio-visual MMKG for precise retrieval, it induces concept-level, single-hop graphs that rarely capture temporal or causal dependencies.
Therefore, constructing a multi-hop, modality-aware knowledge source across audio and visual streams is essential for query-relevant retrieval and reliable spatio-temporal reasoning.

\textbf{Second}, most multimodal retrieval strategies~\cite{jeong-etal-2025-videorag, ghosh2024recap, park2025vat, yeo2025universalrag} rely on similarity search in shared embedding spaces, which captures the query’s broad semantics but misses fine-grained cues.
They often fail to select fine-grained, query-relevant knowledge, retrieve off-topic content, and add redundant context.
Even when retrieved knowledge aligns with the query, facts that do not contribute to the answer introduce noise in the MLLM context.
As shown in \cref{fig:intro}-(b), such simple RAG frameworks, even when they retrieve context-matched knowledge for audio-visual queries, inject noisy evidence that fails to improve response.


To address these limitations, we propose M$^3$KG-RAG, an end-to-end, graph-enhanced RAG framework that constructs a multi-hop MMKG for modality-wise retrieval and supplies only query-aligned, answer-supportive knowledge.
Specifically, we transform raw multimodal corpora into a multi-hop MMKG (M$^3$KG) with a lightweight, collaborative multi-agent pipeline in three steps.
First, we perform \textbf{(i) Context-Enriched Triplet Extraction}, which captures knowledge-intensive entities and relations containing temporal and cross-modal cues.
As triplets alone lack enough context for reliable reasoning~\cite{zha2024m2conceptbase, park2025vat}, we perform \textbf{(ii) Knowledge Grounding} to obtain canonical entity identifiers and descriptions using external resources and tools.
Finally, \textbf{(iii) Context-Aware Description Refinement} aligns entity descriptions with the surrounding multimodal context to ensure consistency and specificity. 
In addition, we incorporate a \textbf{Self-Reflection Loop} to prevent possible hallucinated or misaligned descriptions during construction.

Additionally, we introduce \textit{Grounded Retrieval And Selective Pruning (GRASP)} to keep only query-relevant and answer-useful subgraphs. GRASP first leverages off-the-shelf multimodal grounding models~\cite{liu2024grounding, xu2024towards} to drop triplets not appearing in the query, and then applies a light LLM~\cite{team2024qwen2} to prune triplets that do not contribute to answering the question.
As shown in \cref{fig:intro}-(c), our framework retrieves knowledge tightly linked to the query from the constructed multi-hop MMKG and passes only answer-relevant evidence to the MLLMs.
Extensive experiments across diverse audio, video, and audio-visual QA demonstrate that M$^{3}$KG-RAG achieves substantial performance gains over existing methods. Our contributions are summarized as follows:

\begin{itemize}
    
    \item We present M$^{3}$KG-RAG, an end-to-end framework that integrates a multi-hop MMKG with RAG to enhance audio-visual reasoning in MLLMs.


    \item We propose a three-step, multi-agent pipeline that builds a multi-hop MMKG from raw multimodal corpora, enabling scalable, modality-wise retrieval.
    
    \item We introduce Grounded Retrieval And Selective Pruning (GRASP), which discards graph elements absent from the query or unhelpful for answering and retains only query-relevant, answer-useful subgraphs for the MLLMs.

    \item Through extensive evaluations across diverse multimodal benchmarks, we demonstrate that M$^3$KG-RAG consistently outperforms strong RAG baselines.
    
\end{itemize}
\section{Related Work}
\label{sec:related}
\subsection{Multimodal Large Language Model}

Recent advances in large language models (LLMs)~\cite{team2024qwen2, anil2023palm, dubey2024llama, guo2025deepseek, achiam2023gpt, yang2025qwen3} have showcased strong reasoning and generation capabilities within the language domain. 
This progress has extended to multimodal settings (\textit{e.g.}, vision and audio), leading to the emergence of multimodal large language models (MLLMs).
Early MLLMs, such as Flamingo~\cite{alayrac2022flamingo} and BLIP-2~\cite{li2023blip}, focused on vision–language understanding through lightweight cross-modal interfaces built on top of frozen LLM backbones.
Subsequent works broadened both the scale and reasoning capabilities of MLLMs. 
For instance, LLaVA~\cite{liu2023visual} enables general-purpose image–text understanding through visual instruction tuning using a pretrained vision encoder and LLM. Later variants~\cite{li2024llava, li2025llavanextinterleave} further extend the framework to incorporate video inputs.
Parallel efforts in the audio domain have produced MLLMs capable of reasoning over auditory inputs.
SALMONN~\cite{tang2023salmonn} integrates speech and general sound understanding to LLMs through specialized audio encoders. 
More recently, Kimi-Audio~\cite{ding2025kimi} achieves strong results across audio understanding, generation, and conversational tasks as an audio foundation model.
Motivated by the success of both modalities, recent MLLMs focus on enhancing joint audio-visual understanding.
Video-LLaMA2~\cite{cheng2024videollama} realizes this with a dual branch for spatial–temporal video and audio cues, improving event and scene comprehension under synchronized fusion. 
Qwen2.5-Omni~\cite{xu2025qwen2} further targets real-time interaction, supporting perception and generation across multiple modalities in a streaming manner. 
Complementing open-source models, commercial models such as GPT-4o~\cite{hurst2024gpt} extend multimodal I/O to low-latency audio–visual dialogue, reflecting a shift toward tightly integrated perception and reasoning.

\begin{figure*}[!ht]
\centering
\includegraphics[width=\linewidth]{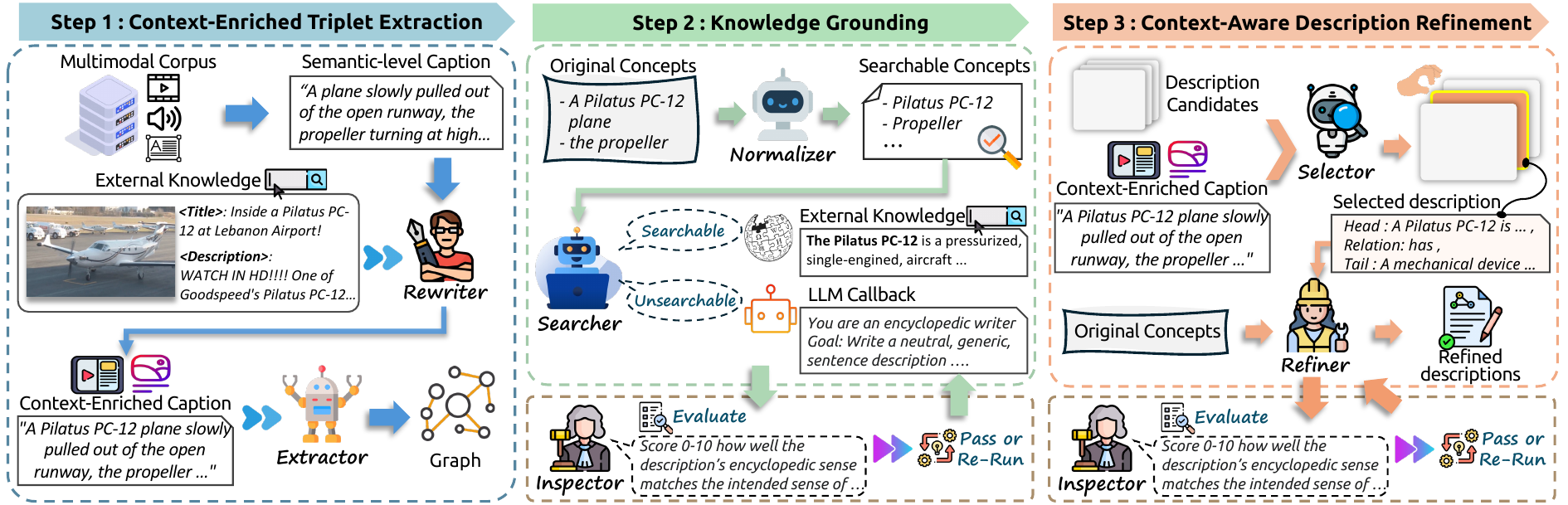}
\caption{
    \textbf{An overview of the M$^3$KG construction pipeline.}
    The pipeline consists of three steps: 
    (i) \textbf{Context-Enriched Triplet Extraction}, which rewrites multimodal captions into knowledge-intensive text and extracts entity–relation triplets; 
    (ii) \textbf{Knowledge Grounding}, linking normalized entities to open knowledge bases to obtain candidate descriptions; 
    (iii) \textbf{Context-Aware Description Refinement}, selecting and rewriting the most context-relevant descriptions for each entity; 
    and \textbf{Self-Reflection Loop}, where an inspector agent validates or re-runs uncertain outputs to ensure graph quality.
    }
\vspace{-1.em}
\label{fig:pipeline}
\end{figure*}

\subsection{Multimodal RAG}

Retrieval-Augmented Generation (RAG) conditions LLM generation on retrieved knowledge, improving grounding and factuality~\cite{lewis2020retrieval, guu2020retrieval, izacard2020leveraging, borgeaud2022improving, gao2023precise, fan2024survey}.
To support multi-hop compositional reasoning across entities and relations, graph-based RAG represents knowledge as entity–relation graphs.
Along this line, GraphRAG~\cite{edge2024local} improves coherence and coverage via graph-aware indexing and community summaries, while LightRAG~\cite{guo2024lightrag} uses dual-level retrieval for efficient, interpretable selection. 
HippoRAG2~\cite{gutierrezrag} strengthens multi-hop retrieval with PPR-style walks and on-the-fly passage integration under tighter online LLM.
With the advent of MLLMs, RAG extends beyond text to multimodal grounding, yet text-only graph RAG methods do not account for intrinsic multimodal semantics, making direct transfer difficult.
In response, multimodal knowledge graph (MMKG) based approaches have emerged, encoding entities and relations across modalities.
MR-MKG~\cite{lee2024multimodal} leverages MMKG with graph encoding and cross-modal alignment to improve multimodal reasoning.
M$^2$ConceptBase~\cite{zha2024m2conceptbase} organizes image–text corpora into a concept-centric MMKG and couples it with graph-aware retrieval, improving grounding and answer accuracy for multimodal QA and description tasks.
Pushing beyond image–text, VAT-KG~\cite{park2025vat} integrates visual, audio, and text into an MMKG and proposes a RAG protocol tailored to audio–visual MLLMs, improving faithfulness under audio–visual queries.
However, its graph construction is largely single-hop, and its RAG framework mainly relies on similarity-based search, which can admit off-topic neighbors and redundant context.
In contrast, we build a multi-hop MMKG and execute fine-grained, query-conditioned retrieval, supplying answer-useful context.

\section{Method}
\label{sec:method}

In this section, we detail the M$^3$KG-RAG paradigm, emphasizing its core architecture and contributions. Sec.~\ref{sec:mmkg_construct} presents our multi-hop multimodal knowledge graph (M$^3$KG) construction pipeline. Sec.~\ref{sec:mmrag_framework} introduces our multimodal RAG framework equipped with the proposed GRASP.

\subsection{\texorpdfstring{M$^3$KG Construction with Multi Agents}{M3KG Construction with Multi Agents}}
\label{sec:mmkg_construct}

Our M$^3$KG-RAG improves the retrieval scheme by constructing a multi-hop MMKG that enables scalable, in-depth reasoning over multiple multimodal triplets. 
However, since constructing a multi-hop MMKG requires multiple stages beyond simple graph connectivity, we design a lightweight, collaborative multi-agent pipeline with our own specialized LLM agents—\textit{rewriter}, \textit{extractor}, \textit{normalizer}, \textit{searcher}, \textit{selector}, \textit{refiner}, and \textit{inspector}—to balance automation and quality control, in line with recent advances in multi-agent LLM systems and self-reflection~\cite{bo2024reflective, tran2025multi}.
The overall construction process is illustrated in \cref{fig:pipeline}, and the detailed role and method for each step are as follows.

\vspace{-1.em}
\paragraph{Step 1: Context-Enriched Triplet Extraction} 
Our multi-hop MMKG construction starts from a raw multimodal corpus $\mathcal{C}=\{(x_n^{text},x_n^{{audio}},x_n^{visual})\}_{n=1}^{N}$ of $N$ aligned samples, where $x_n^{text}, x_n^{audio}, x_n^{visual}$ denote the text, audio, and visual data for sample $n$.
Much of the text in $\mathcal{C}$ is semantically generic~\cite{chen2020vggsound, kim2019audiocaps}, which limits its utility as external knowledge for MLLMs. 
Following prior MMKG construction studies~\cite{park2025vat}, we develop a new \textit{rewriter} that converts semantic-level caption $x_n^{text}$ into a context-enriched caption $\tilde{x}_n^{text}$ by incorporating external knowledge—titles and descriptions collected via a crawler—to supply knowledge-intensive context.
Motivated by recent advances in open information extraction leveraging LLMs~\cite{zhanggraph2025survey, zhang2024extract}, we further introduce an \textit{extractor} that parses $\tilde{x}_n^{text}$ and returns triplets $\mathcal{T}_n = \{(h_i, r_i, t_i)\}_{i=1}^{K_n}$, where $h_i$, $r_i$, and $t_i$ denote the head entity, relation, and tail entity and $K_n=|\mathcal{T}_n|$ varies by input.
As the rewritten $\tilde{x}_n^{text}$ is knowledge-intensive and summarizes the overall multimodal context, the extracted triplets often capture relations among long-tail or uncommon entities—cases that MLLMs commonly miss or misidentify.

\vspace{-1.em}
\paragraph{Step 2: Knowledge Grounding} 
While transforming the corpus into a graph structure enables efficient entity-level access, connections alone offer limited guidance to MLLMs. To enrich the MMKG beyond connectivity, we ground encyclopedic descriptions to entities in this step.
Head and tail entities in $\mathcal T$ often include modifiers or variant surface forms that hinder look-up (\eg, “small brown dog” vs. “dog”). Thus, the \textit{normalizer} first maps each entity mention to a canonical, searchable concept by removing non-essential modifiers, preserving the source word order when appropriate, and standardizing to a singular noun phrase.
Given the normalized concepts, the \textit{searcher} queries open knowledge bases (\eg, Wikipedia, Wiktionary) and uses a crawler to retrieve a compact set of candidate descriptions for each entity. 
Since open knowledge bases cannot cover every textual concept, we include a lightweight LLM callback to fill missing descriptions. Subsequently, we have a candidate description set $\mathcal{D}$ for every normalized concept.

\begin{figure*}[t]
    \centering
    \includegraphics[width=\linewidth]{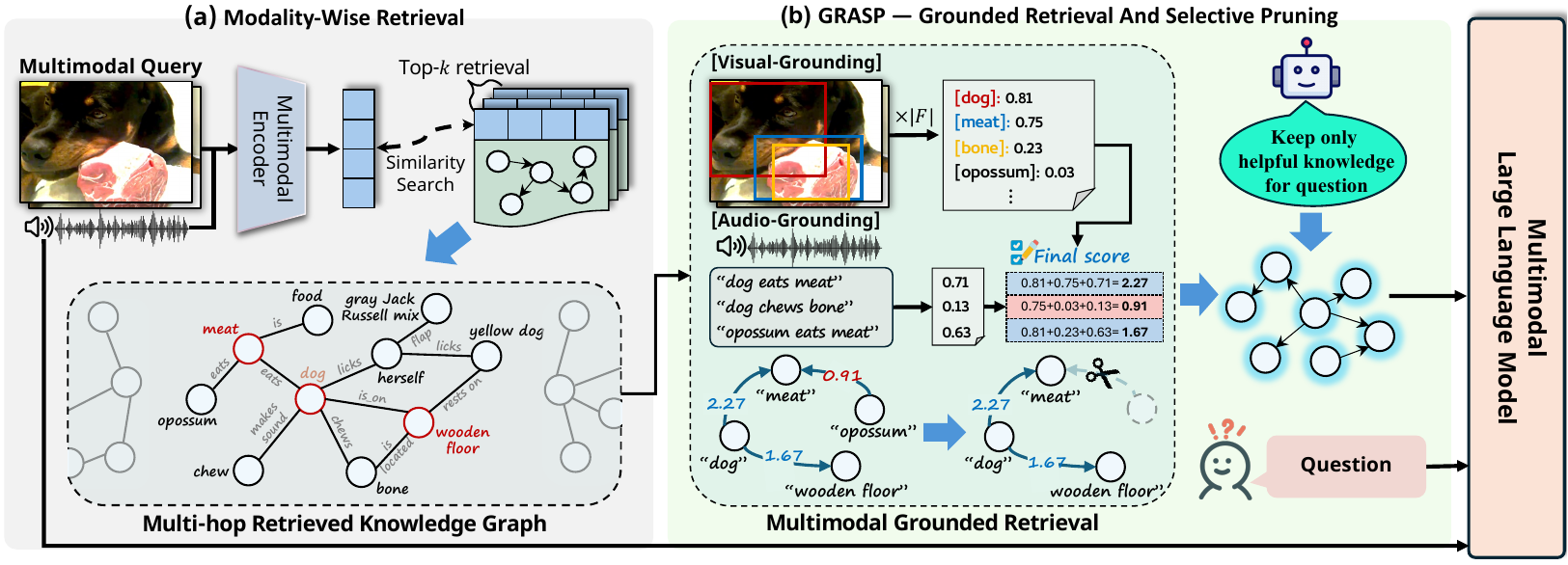}
    \caption{
    \textbf{Overview of the Multimodal RAG framework.}
    %
    The framework consists of two components:
    \textbf{(a) Modality-Wise Retrieval}, which retrieves multi-hop triplets aligned with the query from the M$^3$KG; and
    \textbf{(b) GRASP (Grounded Retrieval And Selective Pruning)}, which uses visual and/or audio grounding models to check entity presence and prunes triplets that are off-topic or non-informative.
    The resulting subgraph is then provided to an MLLM for query-relevant, evidence-grounded audio-visual reasoning.}
    \vspace{-1.em}
    \label{fig:rag_pipeline}
\end{figure*}

\vspace{-1.em}
\paragraph{Step 3: Context-Aware Description Refinement}
A single term can carry multiple meanings, (\eg, “bank”: financial institution vs. a river bank). To make entities accurately informative, the \textit{selector} chooses the most context-appropriate description from the candidate set, using the context-enriched caption from \textbf{Step 1} as guidance for each normalized concept. This keeps descriptions aligned with the context and filters out off-topic ones.
Since the selected descriptions are written for a normalized concept rather than the original heads and tails in $\mathcal{T}$, the \textit{refiner} adapts the chosen description to the original concept’s phrasing to inject the original semantics while preserving the selected content. After this step, we obtain a refined description set $\hat{\mathcal{D}}$ for all entities.

\vspace{-1.em}
\paragraph{Self-Reflection Loop}
To ensure knowledge graph quality, we introduce \textit{Inspector} that implements a self-reflection loop within our construction pipeline.
When the task extends beyond simple information extractions and instead relies on the language model's implicit knowledge (\eg, Step 2 LLM Callback or Step 3 Rewriter), errors may occur. Accordingly, the inspector reviews these outputs and either passes them or returns a re-run signal to the producing agent.

\vspace{-1.em}
\paragraph{Resulting M$^3$KG}
Starting from the text data $x_n^{text}$ in multimodal corpus $\mathcal{C}$, our multi-agent pipeline with a self-reflection loop constructs a multi-hop knowledge graph and links its triplets to the corresponding audio-visual data ($x_n^{audio}, x_n^{visual} $), yielding the following multi-hop multimodal knowledge graph:

\vspace{-0.5em}
\begin{equation}
\label{eq:graph}
    \mathcal{G} = \{\mathcal{E}, \mathcal{R}, \mathcal{T}, \hat{\mathcal{D}}, \mathcal{A}, \mathcal{V}, \mathcal{L}\},
\end{equation}
where\ 
$\mathcal{E}$ is the set of entities;\ 
$\mathcal{R}$ the set of relations;\ 
$\mathcal{T}\subseteq \mathcal{E}\times\mathcal{R}\times\mathcal{E}$ the set of triplets;\ 
$\hat{\mathcal{D}}=\{\,d_e\,\}_{e\in\mathcal{E}}$ the per-entity refined descriptions;\ 
$\mathcal{A}=\{x_n^{\mathrm{audio}}\}_{n=1}^{N}$ and $\mathcal{V}=\{x_n^{\mathrm{visual}}\}_{n=1}^{N}$ the audio/visual items;\ 
$\mathcal{L}\subseteq \mathcal{T}\times(\mathcal{A}\cup\mathcal{V})$ the links from triplets to associated audio/visual data. 
The resulting multi-hop MMKG satisfies the following coverage property:
\begin{equation}
\forall\, \mathfrak{t}\in\mathcal{T},\ \exists\, x\in(\mathcal{A}\cup\mathcal{V})\ \text{s.t.}\ (\mathfrak{t},x)\in\mathcal{L}.
\label{eq:graph_link}
\end{equation}

Consequently, since every triplet links to at least one multimodal item, all facts are eligible for retrieval under multimodal queries, providing full graph coverage.

\subsection{Multimodal RAG Framework}
\label{sec:mmrag_framework}
To deliver query-relevant and answer-useful context only, we design a multimodal RAG framework composed of (i) Modality-Wise Retrieval over the MMKG to gather candidates aligned with the input modalities, and (ii) GRASP to keep only knowledge that is relevant to the multimodal query and useful for answering the question.
An overview of the multimodal RAG framework is shown in Fig.~\ref{fig:rag_pipeline}.

\vspace{-1.em}
\paragraph{Modality-Wise Retrieval} 
Shared embedding spaces from multimodal encoders often exhibit a modality gap where cross-modal distances are not comparably calibrated~\cite{yeo2025universalrag}. 
Consequently, querying a knowledge base indexed in a different modality—for instance, a video query against text embeddings—often yields off-topic neighbors.
To bridge the modality gap, we find the items of the same modality as the query in $\mathcal{G}$ and lift them to triplets. This procedure is enabled by Eq.~\ref{eq:graph_link}, which guarantees that each triplet in $\mathcal G$ is linked to at least one audio or visual item. 
Concretely, we obtain query embeddings with multimodal foundation models (\eg, InternVL2~\cite{wang2024internvideo2} for video and CLAP~\cite{wu2023large} for audio) and search a FAISS~\cite{johnson2019billion} index built over $\mathcal{G}$'s audio/visual items using L2 distance in the embedding space. 
We first retrieve the top-$k$ nearest items and then keep only candidates within a distance threshold $\tau$ of the query to avoid off-topic neighbors.
When both audio and video are provided, we form a simple vector concatenation and apply the same search.
Let $\mathcal {S} \subseteq (\mathcal{A} \cup \mathcal{V})$ denote the set of audio/visual items selected by the above retrieval, we then lift items to triplets to obtain the query-relevant initial graph:
\begin{equation}
\mathcal{G}_{init} = \{\, \mathfrak{t} \in \mathcal{T} \mid \exists\, x \in \mathcal{S},\; (\mathfrak{t},x) \in \mathcal{L} \,\}.
\end{equation}

 


\vspace{-1.5em}
\paragraph{GRASP (Grounded Retrieval And Selective Pruning)}
After mitigating the modality gap, we obtain a query-aligned initial graph $~\mathcal{G}_{init}$. Yet, its similarity-only retrieval can lack fine-grained alignment or include knowledge that may not be useful to answer the question.
For instance, if the question asks \textit{“What instrument is being played?”}, only knowledge about the instruments present is useful, and the remaining triplets mostly become noise for the MLLMs.
To address these limitations, we design a Grounded Retrieval And Selective Pruning (GRASP) to align the graph finely with the query and retain only answer-useful knowledge.
Specifically, we first use off-the-shelf multimodal grounding models to verify whether entities or triplets in $~\mathcal{G}_{init}$ appear in the query’s audio and/or visual streams.
For visual grounding, we use GroundingDINO~\cite{liu2024grounding} on four uniformly sampled frames $F$ from the query video $q_v$, obtaining per-frame detection confidences $\Phi_v(e; f)$ for each entity $e \in \mathcal{G}_{init}$ on each frame $f$. We then take the maximum across the sampled frames as the visual presence score of each entity:
\begin{equation}
s_{v}(e \mid q_v) \;=\; \max_{f \in F}\, \Phi_{v}(e; f).
\vspace{-0.5em}
\end{equation}

We derive a triplet-level visual presence score by summing the presence scores of the head and tail entities and prune triplets whose score falls below $\eta_v$.

For audio grounding, we use a Text-to-Audio Grounding model (TAG)~\cite{xu2024towards}. Unlike visual signals, audio is not as easily factorized into independent snapshots. Accordingly, we convert each triplet $\mathfrak{t}$ into a natural sentence $\sigma(\mathfrak{t})$ (\eg, "h r t") and use the TAG scoring function $\Phi_{a}$ to measure how strongly $\sigma(\mathfrak{t})$ is grounded in the query audio $q_a$, resulting in the audio presence score:

\vspace{-0.5em}
\begin{equation}
    s_{a}(\mathfrak{t} \mid q_a) = \Phi_{a}(\sigma(\mathfrak{t}); q_{a}).
\end{equation}

Similar to the visual case, we drop triplets whose audio presence score falls below $\eta_a$.
When both audio and visual streams are available, we simply sum their presence scores and remove triplets whose fused score is below $\eta_{av}$. This grounding step yields the grounded subgraph $\mathcal{G}_{grd}$.

After obtaining the grounded subgraph $\mathcal{G}_{grd}$, we remove knowledge that is not helpful for answering the question. A lightweight LLM~\cite{xu2025qwen3} produces a binary mask over triplets under a conservative keep-or-drop policy, yielding $\mathcal{G}_{\mathrm{GRASP}}$. This procedure filters unhelpful and off-topic triplets while keeping answer-supportive knowledge.

\vspace{-1.em}
\paragraph{Graph-Augmented Generation}
Our M$^3$KG-RAG targets MLLMs that jointly reason over video, audio, and text.
Given the retrieved subgraph $\mathcal{G}_{\mathrm{GRASP}}$, we condition the MLLM by concatenating the multimodal query $q$ with the graph context. For each triplet $(h,r,t)\in\mathcal{G}_{\mathrm{GRASP}}$, we include the relation $r$ together with the entity–description pairs $\langle h, d_h \rangle$ and $\langle t, d_t \rangle$, where $d_h$ and $d_t$ are the refined descriptions of $h$ and $t$, respectively. 
Formally, our graph-enhanced generation is defined as follows:

\vspace{-1.em}
\begin{equation}
\label{eq:prompt_aug}
p_{\mathrm{aug}}
=
q \,\Vert\,
\left(
\bigcup_{(h,r,t)\in\mathcal{G}_{\mathrm{GRASP}}}
\langle h, d_h\rangle \stackrel{r}{\longrightarrow} \langle t, d_t\rangle
\right).
\end{equation}

By providing $p_{\mathrm{aug}}$ to the MLLMs, we inject query-relevant, answer-useful knowledge and supply the inter-entity relations together with detailed entity attributions, which improves the MLLM's reasoning capabilities.

\begin{table*}[t]
\small
\begin{center}
    \resizebox{\linewidth}{!}{%
    \begin{tabular}{C{2.5cm}|C{3.0cm}|C{3.0cm}|C{3.0cm}|C{3.0cm}}
        \toprule
        \multirow{2}{*}{\textbf{MLLM}} & \multirow{2}{*}{\textbf{Method}} & \textbf{Audio QA} & \textbf{Video QA} & \textbf{Audio-Visual QA} \\
        \cmidrule(lr){3-5}
        & & AudioCaps-QA & VCGPT & VALOR \\
        \midrule
        \multirow{6}{*}{VideoLLaMA2} & None             & 43.13 & 39.09 & 25.66 \\
                                     & Wikidata         & 43.58 & 38.58 & 26.43 \\
                                     & VTKG             & 43.02 & 38.88 & 25.92 \\
                                     & M$^2$ConceptBase & 42.19 & 39.31 & 25.93 \\
                                     & VAT-KG    & 44.60 & 39.42 & 28.30 \\
                                     & M$^3$KG-RAG & \textbf{53.23} & \textbf{39.92} & \textbf{29.25} \\
        \midrule
        \multirow{6}{*}{Qwen2.5-Omni} & None             & 49.00 & 42.21 & 32.42 \\
                                      & Wikidata         & 49.78 & 40.82 & 30.28 \\ 
                                      & VTKG             & 48.95 & 42.96 & 32.70 \\
                                      & M$^2$ConceptBase & 49.78 & 42.78 & 32.31 \\
                                      & VAT-KG    & 51.30 & 43.50 & 35.44 \\
                                      & M$^3$KG-RAG & \textbf{60.77} & \textbf{44.35} & \textbf{44.67} \\
        \bottomrule
    \end{tabular}}
    \caption{\textbf{Overall performance.} We report Model-as-Judge (M.J.) scores (higher is better). Across all benchmarks and for both MLLMs, M$^3$KG-RAG provides the largest and most consistent gains over the no-retrieval and MMKG-based baselines. The best results are \textbf{bolded}.}
    \label{tab:exp_1}
    \vspace{-2.em}
\end{center}
\end{table*}


\section{Experiments}
\label{sec:exp}

\subsection{Experimental Setup}
\label{sec:exp_setup}

\paragraph{Datasets}
To highlight the diverse modality coverage of M$^3$KG-RAG, we evaluate on three multimodal tasks: Audio QA, Video QA, and Audio-Visual QA.
For Audio-QA, we use AudioCaps-QA~\cite{wang2025audiobench}, which provides human-annotated QA pairs built on AudioCaps corpus~\cite{kim2019audiocaps}.
For Video-QA, we adopt the VideoChatGPT (VCGPT) benchmark~\cite{maaz2024video}, which is built on videos curated from ActivityNet~\cite{caba2015activitynet}.
For Audio-Visual QA, we evaluate on the VALOR~\cite{liu2024valor} benchmark, which explicitly requires joint reasoning over synchronized audio and visual streams for response.

\vspace{-1.em}
\paragraph{MLLMs}
We apply our M$^3$KG-RAG to three MLLMs capable of joint audio-visual understanding to assess its effectiveness.
Specifically, we employ VideoLLaMA2~\cite{cheng2024videollama} and Qwen2.5-Omni~\cite{xu2025qwen2}, both advanced open-source models that jointly process audio and visual streams and serve as strong baselines for audio–visual reasoning.
We further adopt GPT-4o~\cite{hurst2024gpt}, a strong commercial model with substantial implicit knowledge and capacity, to examine whether our method remains effective even for such high-capacity models.

\vspace{-1.em}
\paragraph{Baseline Methods}
Following prior multimodal RAG work~\cite{park2025vat}, we compare M$^3$KG-RAG against five baselines: (i) None—the MLLM answers without external knowledge; (ii) Wikidata~\cite{wang2021kepler} + naïve RAG with retrieval in a shared embedding space~\cite{wang2024internvideo2,wu2023large} between the text KG and the multimodal query; (iii) VTKG~\cite{lee2023vista} + naïve RAG (image–text MMKG), matching visual queries to images in MMKG via a vision–language space (\eg, CLIP~\cite{radford2021learning}) and audio queries in the CLAP~\cite{wu2023large} audio–text space; (iv) M$^2$ConceptBase~\cite{zha2024m2conceptbase}, following VTKG’s protocol; and (v) VAT-KG~\cite{park2025vat}, evaluated under its released RAG protocol that accounts for audio–visual streams.

\vspace{-1.em}
\paragraph{Implementation Details}
We implement the multi-hop MMKG construction pipeline with a lightweight multi-agent stack built on a single backbone LLM (Qwen3-8B)~\cite{xu2025qwen3}, using only the training splits of our evaluation benchmarks.
In modality-wise retrieval, we set $k=5$ and select the top-$k$ best-matching items per query, then expand each to its connected multi-hop subgraph.
Since our benchmarks span different audio–visual distributions, we set the modality-wise distance threshold $\tau$ and GRASP presence threshold $\eta$ separately per benchmark as follows: for AudioCaps-QA, $\tau=3.0$ and $\eta_a=0.5$; for VCGPT, $\tau=0.15$ and $\eta_v=1.5$; and for VALOR, $\tau=4.5$ and $\eta_{av}=1.2$. These values are held constant within each benchmark across all experiments. 
All experiments use a single NVIDIA H100 GPU.
Note that further details are provided in the supplementary material.

\vspace{-1.em}
\paragraph{Evaluation Metrics}
We implement two evaluation schemes. First, since our benchmarks consist of open-ended QA with free-form responses, we adopt an off-the-shelf Model-as-Judge (M.J.) protocol~\cite{wang2025audiobench}, where an LLM judge~\cite{grattafiori2024llama} scores each response.
Second, in line with established RAG evaluation~\cite{edge2024local, guo2024lightrag, ren2025videorag}, we report a win-rate preference protocol where the LLM judge compares the two responses (ours vs. a baseline) and selects the preferred response based on multiple criteria. 
We make it reference-aware by providing the judge with the reference answer during comparison, which reduces verbosity bias and yields a more reliable, multi-dimensional assessment.

\subsection{Quantitative Results}
\label{sec:exp_quant}
We compare M$^3$KG-RAG against text-KG and multimodal-KG baselines on Audio-QA, Video-QA, and Audio-Visual QA. The overall results are summarized in Table~\ref{tab:exp_1}. Across all benchmarks, M$^3$KG-RAG yields significant gains over the base MLLMs, indicating that modality-wise retrieval and GRASP deliver knowledge that is both tightly aligned with the query and directly useful for answering. 
In contrast, other baselines tend not to consistently improve the MLLMs.
Specifically, text KG with naïve RAG (Wikidata) yields weak or even negative deltas, as retrieval ignores the temporal nature of audio–visual queries, often retrieving off-context neighbors and injecting noisy facts that do not support the answer. Image-text KGs with naïve RAG (VTKG, M$^2$ConceptBase) partially account for visual cues via images in MMKGs but still miss query dynamics, leading to limited impact on response quality and occasional degradation.
VAT-KG, which considers audio–visual streams, improves all baselines uniformly.
However, its largely single-hop MMKG captures only local, concept-level facts. The MLLM therefore receives only shallow, fragmentary context, so the knowledge implicitly encoded in the underlying multimodal data is only partially exploited, and performance gains remain mostly marginal. 
In contrast, M$^3$KG-RAG builds multi-hop neighborhoods that aggregate temporally and semantically related evidence across modalities and, together with modality-wise retrieval and GRASP, delivers query-focused, answer-supporting knowledge that more faithfully reflects the multimodal query.
Consequently, M$^3$KG-RAG achieves notable gains over VAT-KG on every benchmark.
These findings become more pronounced with a stronger commercial MLLM. 
As shown in Table~\ref{tab:gpt}, even with substantial built-in knowledge, GPT-4o paired with M$^3$KG-RAG improves across all benchmarks and exhibits larger gains than with VAT-KG, reinforcing that multi-hop evidence together with GRASP provides a diverse, answer-supporting context that the model can exploit more effectively.

\begin{table}[t]
\centering
\vspace{-0.1in}
\small
\setlength{\tabcolsep}{2.6pt}
\renewcommand{\arraystretch}{0.95}
\resizebox{\linewidth}{!}{
\begin{tabular}{@{}lcccccc@{}}
\toprule
\textbf{} & \multicolumn{2}{c}{\textbf{AudioCaps-QA}} & \multicolumn{2}{c}{\textbf{VCGPT}} & \multicolumn{2}{c}{\textbf{VALOR}} \\
\cmidrule(lr){2-3} \cmidrule(lr){4-5} \cmidrule(lr){6-7}
& Baseline & \textbf{Ours} & Baseline & \textbf{Ours} & Baseline & \textbf{Ours} \\
\midrule
\multicolumn{7}{l}{\textit{Baseline: None}} \\
\midrule
Comprehensiveness &  15.9\% & \textbf{84.1\%} &  47.6\% & \textbf{52.4\%\ } &  39.8\% & \textbf{60.2\%\ } \\
Diversity         &  20.3\% & \textbf{79.7\%} &  37.8\% & \textbf{62.2\%\ } &  45.5\% & \textbf{54.5\%\ } \\
Empowerment       &  14.0\% & \textbf{86.0\%} &  42.1\% & \textbf{57.9\%\ } &  40.1\% & \textbf{59.9\%\ } \\
Overall           &  15.2\% & \textbf{84.8\%} &  47.0\% & \textbf{53.0\%\ } &  39.8\% & \textbf{60.2\%\ } \\
\cmidrule(lr){1-7}
\multicolumn{7}{l}{\textit{Baseline: Wikidata}} \\
\midrule
Comprehensiveness &  14.9\% & \textbf{85.1\%} &  48.3\% & \textbf{51.7\%\ } &  40.3\% & \textbf{59.7\%\ } \\
Diversity         &  22.4\% & \textbf{77.6\%} &  47.4\% & \textbf{52.6\%\ } &  \textbf{55.5\%} & 44.5\%\  \\
Empowerment       &  12.0\% & \textbf{88.0\%} &  39.6\% & \textbf{60.4\%\ } &  40.8\% & \textbf{59.2\%\ } \\
Overall           &  13.7\% & \textbf{86.3\%} &  44.5\% & \textbf{55.5\%\ } &  40.8\% & \textbf{59.2\%\ } \\
\cmidrule(lr){1-7}
\multicolumn{7}{l}{\textit{Baseline: VTKG}} \\
\midrule
Comprehensiveness &  20.8\% & \textbf{79.2\%} &  49.1\% & \textbf{50.9\%\ } &  39.1\% & \textbf{60.9\%\ } \\
Diversity         &  33.8\% & \textbf{66.2\%} &  45.9\% & \textbf{54.1\%\ } &  45.2\% & \textbf{54.8\%\ } \\
Empowerment       &  21.2\% & \textbf{78.8\%} &  46.6\% & \textbf{53.4\%\ } &  39.2\% & \textbf{60.8\%\ } \\
Overall           &  21.2\% & \textbf{78.8\%} &  49.1\% & \textbf{50.9\%\ } &  39.4\% & \textbf{60.6\%\ } \\
\cmidrule(lr){1-7}
\multicolumn{7}{l}{\textit{Baseline: M$^2$ConceptBase}} \\
\midrule
Comprehensiveness &  21.2\% & \textbf{78.8\%} &  41.8\% & \textbf{58.2\%\ } &  38.2\% & \textbf{61.8\%\ } \\
Diversity         &  28.3\% & \textbf{71.7\%} &  43.9\% & \textbf{56.1\%\ } &  45.1\% & \textbf{54.9\%\ } \\
Empowerment       &  19.7\% & \textbf{80.3\%} &  44.6\% & \textbf{55.4\%\ } &  38.6\% & \textbf{61.4\%\ } \\
Overall           &  21.0\% & \textbf{79.0\%} &  44.3\% & \textbf{55.7\%\ } &  38.3\% & \textbf{61.7\%\ } \\
\cmidrule(lr){1-7}
\multicolumn{7}{l}{\textit{Baseline: VAT-KG}} \\
\midrule
Comprehensiveness &  26.1\% & \textbf{73.9\%} &  48.4\% & \textbf{51.6\%\ } &  41.4\% & \textbf{58.6\%\ } \\
Diversity         &  34.8\% & \textbf{65.2\%} &  46.6\% & \textbf{53.4\%\ } &  48.3\% & \textbf{51.7\%\ } \\
Empowerment       &  24.3\% & \textbf{75.7\%} &  43.5\% & \textbf{56.5\%\ } &  42.1\% & \textbf{57.9\%\ } \\
Overall           &  25.6\% & \textbf{74.4\%} &  47.6\% & \textbf{52.4\%\ } &  41.8\% & \textbf{58.2\%\ } \\
\bottomrule
\end{tabular}
}
\caption{\textbf{Win-rate comparison.} Pairwise win rates (\%) of each baseline versus M$^3$KG-RAG across three benchmarks and four criteria. Columns show the preference rate of the \textit{Baseline} and \textit{Ours}, with the higher win rate in each pair highlighted in \textbf{bold}.}
\label{tab:winlose_ours}
\vspace{-2.em}
\end{table}

Win-rate preference results in Table~\ref{tab:winlose_ours} corroborate the M.J. scores, showing consistent preference for M$^3$KG-RAG over baselines across benchmarks and criteria.
We observe higher \textbf{Comprehensiveness} because richer multi-hop evidence aggregates the key entities and relations needed to answer the query end-to-end, aided by refined entity descriptions for clarity.
\textbf{Diversity} improves as the multi-hop MMKG offers several distinct evidence chains, while pruning removes off-topic or duplicate content.
\textbf{Empowerment} benefits from a strictly query-relevant context that reduces hallucination and steers the model toward concrete, answer-supporting details rather than generic filler.
Together, these effects yield stronger \textbf{Overall} preferences in pairwise comparisons.


\begin{table}[ht]
\vspace{-0.8em}
\centering
\label{tab:kg_mllm_results}
\scriptsize
\setlength{\tabcolsep}{8pt}        
\begin{tabular}{c c c c c}   
\toprule
\textbf{MLLM} & \textbf{Method} & \textbf{AudioCaps-QA} & \textbf{VCGPT} & \textbf{VALOR} \\
\midrule
GPT-4o & None               & 56.74 & 49.68 & 46.02 \\
GPT-4o & VAT-KG      & 57.70 & 51.49 & 55.86 \\
GPT-4o & M$^3$KG-RAG & \textbf{59.17} & \textbf{53.05} & \textbf{56.53}    \\
\bottomrule
\end{tabular}
\caption{\textbf{Performance on Commercial MLLM (GPT-4o).} We report M.J. scores (higher is better). The best result is \textbf{bolded}.}
\label{tab:gpt}
\vspace{-1.em}
\end{table}

\begin{table}[ht]
\centering
\setlength{\tabcolsep}{10pt}
\renewcommand{\arraystretch}{1.35}
\label{tab:valor_grasp_min}
\scriptsize
\begin{tabular}{c|cc|c}
\toprule
\multirow{2}{*}{\textbf{MLLM}} & \multicolumn{2}{c|}{\textbf{Method}} & \multirow{2}{*}{\textbf{M.J$\uparrow$ }} \\
\cmidrule(lr){2-3}
 & \makecell{\textbf{Modality-Wise}\\\textbf{Retrieval}} & \textbf{GRASP} & \\
\midrule
\multirow{4}{*}{\textbf{Qwen2.5-Omni}} 
 & \xmark & \xmark & 36.62 \\
 & \cmark & \xmark & 40.91 \\
 & \xmark & \cmark & 36.96 \\
 & \cmark & \cmark & \textbf{44.67} \\
\bottomrule
\end{tabular}
\caption{\textbf{Ablation on VALOR.} Checkmarks denote enabled components. We report the M.J. score; combining Modality-Wise Retrieval and GRASP gives the best score. The best result is \textbf{bolded}.}
\label{tab:abl}
\vspace{-2.em}
\end{table}

\subsection{Ablation Study}
\label{sec:exp_abl}
To explore the effectiveness of our design, we conduct ablations of modality-wise retrieval and GRASP on the VALOR, which requires joint audio–visual reasoning, using Qwen2.5-Omni~\cite{xu2025qwen2} as the base MLLM.
For modality-wise retrieval, we remove the cross-modal links $\mathcal {L}$ that connect multimodal items ($\mathcal{A}$, $\mathcal{V}$) to the triplet set $\mathcal T$ in the M$^3$KG, yielding a text-only KG.
We then convert each triplet $\mathfrak{t}$ into a natural sentence $\sigma(\mathfrak{t})$ and index them using the text encoder of the multimodal embedding model~\cite{wang2024internvideo2, wu2023large}, enabling retrieval for audio-visual queries in a shared embedding space.

As shown in Table~\ref{tab:abl}, using modality-wise retrieval alone keeps retrieval within the query's modality and reduces mismatched evidence. However, relying solely on similarity search cannot verify entity-level relevance to the query or ensure that retrieved evidence supports the answer, leading to limited performance gains.
Using GRASP improves faithfulness via fine-grained pruning, yet the initial graph retrieved from a text-only KG in a shared space is weakly aligned with the audio–visual cues, yielding only modest improvements.

Combining both offers the largest gain: modality-wise retrieval supplies candidates aligned with the query’s audio and visual streams, and GRASP retains only triplets that directly support the question and removes redundancy. Taken together, modality-wise retrieval and grounded pruning are complementary and jointly necessary for multimodal RAG.

\begin{figure*}[ht]
    \centering
    \includegraphics[width=\linewidth]{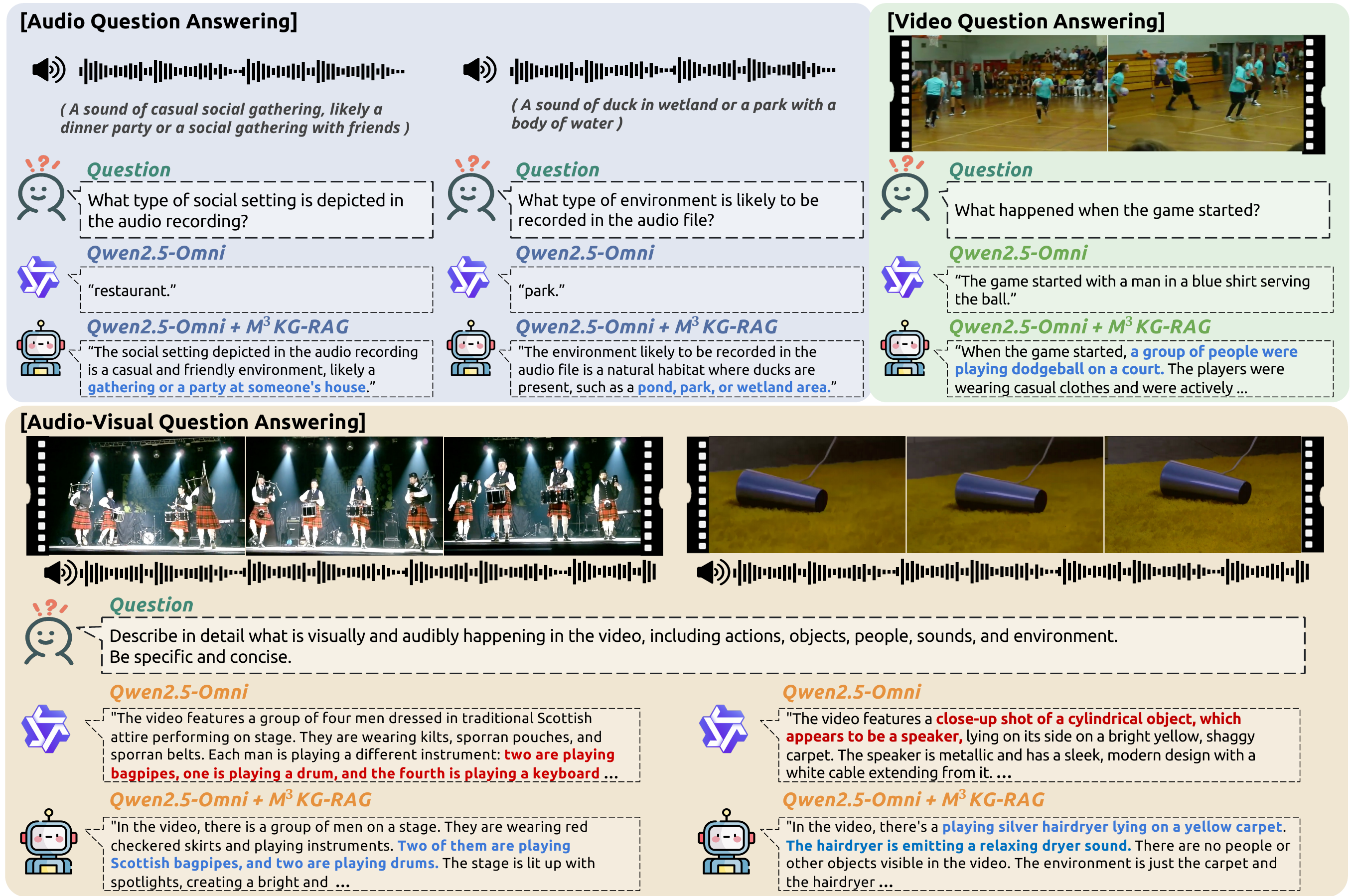}
    \caption{
    \textbf{Qualitative results on various Question Answering tasks.}
    Incorrect and insufficient model responses are highlighted in \textcolor{red}{red}, while correct and sufficient responses are highlighted in \textcolor{blue}{blue}.
    }
    \vspace{-1.em}
    \label{fig:qualitative_result}
\end{figure*}

\subsection{Qualitative Results}
\label{sec:exp_quali}
\cref{fig:qualitative_result} presents qualitative results across Audio, Video, and Audio–Visual QA. With M$^3$KG-RAG, the MLLM produces more specific, context-faithful answers by grounding generation in multi-hop evidence and concise entity descriptions from our modality-wise retrieval and GRASP.

In the case of Audio-QA, the context supplied by M$^3$KG-RAG directly supports answering. For example, for the question “What type of social setting is depicted in the audio recording?” the base model responds “restaurant,” which is loosely related yet misaligned with the asked social setting and lacks sufficient detail. With M$^3$KG-RAG, query-conditioned retrieval selects the correct context, and GRASP passes only answer-useful cues, producing “a gathering or a party at someone’s house.” Likewise, for the environment clip with water, the base MLLM overlooks water-related acoustic attributes, whereas our retrieval highlights duck calls and water ambience, producing “a natural habitat with ducks, such as a pond, park, or wetland.”

In the case of Video-QA, the graph-enhanced context enables precise answering. While the base model misses the action occurring in the video, our RAG conditions retrieval on the visual stream, identifies the scene as dodgeball, and supplies action-aware context that enables a precise answer.

Retrieval conditioned on both audio and visual sharpens predictions in Audio-Visual QA.
For the stage-performance clip, the model hallucinates a keyboard player. With M$^3$KG-RAG, based on the retrieved query-related context, the model produces correct stage context and instrument roles (two bagpipes, two drums).
For the hair-dryer clip, the model tags a "cylinder object" and misclassifies it as a speaker. On the other hand, M$^3$KG-RAG conditions retrieval on the audio-visual query and passes context that links the buzzing audio to a running hair dryer, enabling a precise response.
These results demonstrate that modality-conditioned retrieval with GRASP supplies on-topic, fine-grained context that improves the specificity and faithfulness of answers.
\section{Conclusion}
\label{sec:conclusion}
We introduce M$^{3}$KG-RAG, a novel graph-augmented multimodal RAG framework for enhancing audio–visual reasoning in MLLMs.
Our lightweight multi-agent pipeline constructs a multi-hop multimodal knowledge graph that supports precise modality-wise retrieval and robust knowledge grounding.
Furthermore, the proposed GRASP (Grounded Retrieval And Selective Pruning) scores triplets with visual and audio foundation models and retains only query-relevant, answer-supporting evidence.
Extensive evaluation across diverse multimodal benchmarks highlights the effectiveness of M$^{3}$KG-RAG, with consistent performance gains over strong baselines.
We believe M$^{3}$KG-RAG will serve as a practical foundation for future research in multimodal RAG.

\section*{Acknowledgement}
This work was supported by 
Korea Research Institute for Defense Technology Planning and Advancement (KRIT) - Grant funded by the Defense Acquisition Program Administration (DAPA) (KRIT-CT-23-021, 97\%),
Culture, Sports and Tourism R\&D Program through the Korea Creative Content Agency grant funded by the Ministry of Culture, Sports and Tourism (International Collaborative Research and Global Talent Development for the Development of Copyright Management and Protection Technologies for Generative AI, RS-2024-00345025, 1\%),
the National Research Foundation of Korea(NRF) grant funded by the Korea government(MSIT)(RS-2025-00521602, 1\%),
Institute of Information \& communications Technology Planning \& Evaluation (IITP) \& ITRC(Information Technology Research Center) grant funded by the Korea government(MSIT) (No.RS-2019-II190079, Artificial Intelligence Graduate School Program(Korea University), 1\%), 
and
the NVIDIA Academic Grant Program using NVIDIA RTX PRO 6000 Max-Q Workstation Edition.

{
    \small
    \bibliographystyle{ieeenat_fullname}
    \bibliography{main}
}

\clearpage
\maketitlesupplementary
\setcounter{page}{1}
\setcounter{section}{0}
\renewcommand{\thesection}{\Alph{section}}

\newcommand\framework{{M$^3$KG-RAG}}

\section*{Overview}

This supplementary material provides additional implementation details, experimental results, and qualitative analyses for our proposed framework, M$^3$KG-RAG.

\begin{itemize}
    
    \item In \cref{sec:impl}, we describe the implementation of M$^3$KG-RAG in detail, including the construction of the M$^3$KG and the multimodal RAG framework.
    \item In \cref{sec:analysis}, we present additional analyses of our multimodal RAG framework, including hyperparameter sensitivity and ablations over key components with their computational cost.
    
    \item In \cref{sec:qual}, we provide additional qualitative results of M$^3$KG-RAG on multimodal benchmarks, including win-rate evaluations.

    \item In \cref{sec:limit}, we discuss the limitations of M$^3$KG-RAG and potential directions for improving robustness.
    
\end{itemize}

\section{Extended Implementation Details}
\label{sec:impl}

\subsection{\texorpdfstring{M$^3$KG Construction}{M3KG Construction}}
In this section, we describe how M$^3$KG is built using a lightweight multi-agent pipeline, outlining the roles and designs of each agent. We also summarize the multimodal corpora used for construction.

\subsubsection{Multi-Agents}
We provide detailed descriptions and prompt designs for the multi-agent system used in M$^3$KG construction, including \textit{rewriter}, \textit{extractor}, \textit{normalizer}, \textit{searcher}, \textit{selector}, \textit{refiner}, and \textit{inspector}. 

\myparagraph{Rewriter}
The \textit{rewriter} transforms generic textual descriptions from the raw multimodal corpus into knowledge-intensive captions that are more informative for MLLMs. 
For each data point, it leverages the crawled YouTube title and description to inject unfamiliar concepts and background knowledge that are not explicitly captured in the original text caption. 
The detailed prompt design for the \textit{rewriter} agent is provided in Table~\ref{tab:rewriter}.

\myparagraph{Extractor}
The \textit{extractor} takes the rewritten, knowledge-intensive captions produced by the \textit{rewriter} and extracts structured knowledge in the form of triplets. Building on the LLM-based open information extraction prompt used in VAT-KG~\cite{park2025vat},
we design the prompt for the \textit{extractor} agent, as shown in Table~\ref{tab:extractor}.

\myparagraph{Normalizer}
The \textit{normalizer} operates on the head and tail entities in the triplets extracted by the \textit{extractor} and standardizes them into canonical, searchable concepts. The prompt design for the \textit{normalizer} agent is provided in Table~\ref{tab:prompt_normalizer}.

\myparagraph{Searcher} 
The \textit{searcher} takes the normalized entities produced by the \textit{normalizer} and queries external knowledge resources (e.g., Wikipedia, Wiktionary) to obtain encyclopedic descriptions for each concept. If it cannot find a description from these sources, it invokes an LLM callback~\cite{yang2025qwen3} to generate a brief description for the entity, leveraging context-enriched caption. The prompt designs for the LLM callback of the \textit{searcher} agent are provided in Table~\ref{tab:prompt_searcher_llm}.

\myparagraph{Selector}
The \textit{selector} takes multiple candidate descriptions for each entity and uses the context-enriched caption produced by the \textit{rewriter} as context to select the most appropriate description.
The prompt design for the \textit{selector} agent is provided in Table~\ref{tab:prompt_selector}.

\myparagraph{Refiner}
The \textit{refiner} takes the descriptions selected for each entity’s canonical form and refines them to better match the semantics and surface form of the original entity mention, while preserving the underlying factual content. The prompt design for the \textit{refiner} agent is provided in Table~\ref{tab:prompt_refiner}.

\begin{table}[t]
\small
\centering
\begin{tabular}{p{0.95\linewidth}}
\toprule
\textbf{Rewriter agent} \\
\midrule
\texttt{<system prompt>} \\
You refine video captions using the video's Title and Description.

ORIGINAL CAPTION always has priority. If Title/Description are not clearly referring to the SAME scene/object/action, output the ORIGINAL CAPTION exactly.

Allowed edits ONLY when clearly aligned: 
replace generic nouns with specific terms (breed/species/instrument/model/

place/role), or add 1 short factual attribute. Keep the meaning and keep the length roughly similar (±20\%).

Disallowed: inventing new events, numbers, counts, or speculative facts; adding ads/URLs/hashtags.

Keep the original style. Make the merge natural (not a concat).

Output: ONLY the final caption in English (no labels or explanations). \\
\midrule
\texttt{<user prompt>} \\
Title: \{TITLE\} \\
Description: \{DESCRIPTION\} \\
ORIGINAL CAPTION: \{ORIGINAL\_CAPTION\} \\
Output: \\
\bottomrule
\end{tabular}
\caption{Prompt template for the \textit{rewriter} agent.}
\label{tab:rewriter}
\vspace{-1.em}
\end{table}

\myparagraph{Inspector}
The \textit{inspector} serves as a quality-control agent that implements the self-reflection loop in our construction pipeline. For each entity description produced either by the \textit{refiner} or by the LLM callback of the \textit{searcher}, the \textit{inspector} assigns a plausibility score on a $0$--$10$ scale, conditioned on the context-enriched caption from Step~1. 
Descriptions scoring below $7$ are sent back to the corresponding agent for regeneration and re-scoring; we allow at most three such iterations, after which persistently low-scoring descriptions are discarded to avoid injecting low-quality facts into the graph, while higher-scoring ones are accepted.
The prompt design for the \textit{inspector} agent is provided in Table~\ref{tab:prompt_inspector}.

\begin{table}[t]
\small
\centering
\begin{tabular}{p{0.95\linewidth}}
\toprule
\textbf{Extractor agent} \\
\midrule
\texttt{<system prompt>} \\
You are an expert in extracting structured knowledge from text.
Given a video caption, extract all subject-relationship-object triples in the form (h, r, t).

Extract multiple (h, r, t) triples if applicable.

Each triple must be meaningful and correctly represent relationships in the text.

Output format: ONE triple per line as (h, r, t). No extra text or explanation.

Use concise surface forms that appear (or are directly implied) in the caption.

Do not invent entities or facts not supported by the caption.

Language: English only.
 \\
\midrule
\texttt{<user prompt>} \\
Caption: \{CAPTION\} \\
Output: \\
\bottomrule
\end{tabular}
\caption{Prompt template for the \textit{extractor} agent.}
\vspace{-0.5em}
\label{tab:extractor}
\end{table}

\begin{table}[t]
\small
\centering
\begin{tabular}{p{0.95\linewidth}}
\toprule
\textbf{Normalizer agent} \\
\midrule
\texttt{<system prompt>} \\
You output exactly ONE KB-searchable concept noun phrase (Wikipedia-title-like).
Plain text only, no quotes or extra words.

Output MUST be in English. If \texttt{CONCEPT} is not in English, translate the noun phrase into an English Wikipedia-style title; transliterate proper names if needed (do not add extra words).

Otherwise, use ONLY words from \texttt{CONCEPT}; keep order; you may DROP words (no inventions/translation).

If the output noun phrase is in plural form, convert it to its singular form (e.g., ``dogs'' $\rightarrow$ ``dog'', ``empires'' $\rightarrow$ ``empire'').

Must be a NOUN PHRASE; remove wrappers like ``how to'', ``what is'', guides/tips, articles, years.

Prefer inner object NP (e.g., ``history of jazz music'' $\rightarrow$ ``jazz music'').

Prefer canonical/proper names; preserve original casing. \\[4pt]
\midrule
\texttt{<user prompt>} \\
CONCEPT: \{CONCEPT\} \\
Output: \\
\bottomrule
\end{tabular}
\caption{Prompt template for the \textit{normalizer} agent.}
\label{tab:prompt_normalizer}
\vspace{-1.em}
\end{table}

\begin{table}[t]
\small
\centering
\begin{tabular}{p{0.95\linewidth}}
\toprule
\textbf{Searcher agent (LLM callback)} \\
\midrule
\texttt{<system prompt>} \\
You are an encyclopedic writer. 

Write a neutral, generic, 1--2 sentence encyclopedic description of a concept. 

Use the caption ONLY to disambiguate the intended sense (do not describe the scene).

Return ONLY the final description sentences in plain text---no labels, no lists, no quotes, no extra commentary. \\
\midrule
\texttt{<user prompt>} \\
Concept: \{CONCEPT\} \\
Caption (sense disambiguation only): \{CAPTION\} \\
Output: \\
\bottomrule
\end{tabular}
\caption{Prompt template for the LLM callback of the \textit{searcher} agent.}
\label{tab:prompt_searcher_llm}
\end{table}

\begin{table}[t]
\small
\centering
\begin{tabular}{p{0.95\linewidth}}
\toprule
\textbf{Selector agent} \\
\midrule
\texttt{<system prompt>} \\
You are a selector for concept descriptions. 

The CAPTION is from the same video, and the CONCEPT refers to the concept that appears or is mentioned in this video. 

Choose ONE candidate whose encyclopedic sense best matches that concept in this video's caption. 

Use ONLY the CAPTION to resolve meaning (sense); do not import outside facts. 

Return EXACTLY one candidate's text verbatim; do not edit, merge, summarize, quote, or label it. 

If multiple candidates are similarly valid, prefer the most specific non-speculative candidate. 

If none clearly fits, choose the safest generic candidate (least speculative). 

Return ONLY the chosen candidate text (no extra text). \\
\midrule
\texttt{<user prompt>} \\
CONCEPT: \{CONCEPT\} \\
CAPTION: \{CAPTION\} \\
CANDIDATES: \{ENUMERATED\_CANDIDATES\} \\
Output:  \\
\bottomrule
\end{tabular}
\caption{Prompt template for the \textit{selector} agent.}
\label{tab:prompt_selector}
\vspace{-1.5em}
\end{table}

\begin{table}[t]
\small
\centering
\begin{tabular}{p{0.95\linewidth}}
\toprule
\textbf{Refiner agent} \\
\midrule
\texttt{<system prompt>} \\
You are a refiner for concept descriptions.

Adapt the selected description so it fits the original concept phrasing, preserving the original meaning and keeping the content of the selected description as intact as possible (minimal wording changes only---e.g., adjust possessives like ``my/our/their'', determiners, and surface phrasing to align with the original concept). \\[4pt]

Do NOT add, remove, or invent facts beyond what is in the selected description. \\
Keep the meaning unchanged; only adapt phrasing to match the original concept. \\
Concise: 1--2 sentences, plain text \\ (no lists/quotes/markdown/meta). \\
Do NOT output any reasoning. \\
Return ONLY the rewritten description sentences. \\
\midrule
\texttt{<user prompt>} \\
Concept (original phrasing): \{ORIGINAL\_CONCEPT\} \\
Searchable concept (KB term): \{SEARCHABLE\_CONCEPT\} \\
Selected description (about the searchable concept): \\
\{SELECTED\_DESCRIPTION\} \\
Output: \\
\bottomrule
\end{tabular}
\caption{Prompt template for the \textit{refiner} agent.}
\label{tab:prompt_refiner}
\end{table}

\begin{table}[t]
\small
\centering
\begin{tabular}{p{0.95\linewidth}}
\toprule
\textbf{Inspector agent} \\
\midrule
\texttt{<system prompt>} \\
You are a judge that scores how well an encyclopedic DESCRIPTION matches the intended sense of a CONCEPT. \\
Sense scoring only. \\
Score 0--10 how well the DESCRIPTION's encyclopedic sense matches the intended sense of the CONCEPT (0 = different/irrelevant sense, 10 = perfect sense match). \\
Output a single integer 0--10 with no extra text. \\
\midrule
\texttt{<user prompt>} \\
CONCEPT: \{CONCEPT\} \\
DESCRIPTION: \{DESCRIPTION\} \\
OUTPUT: \\
\bottomrule
\end{tabular}
\caption{Prompt template for the \textit{inspector} agent.}
\label{tab:prompt_inspector}
\vspace{-1.5em}
\end{table}

\subsubsection{\texorpdfstring{Corpora for M$^3$KG Construction}{Corpora for M3KG Construction}}
We construct M$^3$KG from the training splits of three multimodal corpora: AudioCaps~\cite{kim2019audiocaps}, ActivityNet~\cite{caba2015activitynet}, and VALOR~\cite{liu2024valor}. For graph construction, we use only the raw audio-visual content and their associated captions, without accessing any QA annotations.

\myparagraph{AudioCaps}
AudioCaps is an audio captioning corpus built on 10-second clips from AudioSet~\cite{audioset} YouTube videos, where each clip is paired with human-written natural language descriptions of the acoustic scene and salient sound events.

\myparagraph{ActivityNet}
ActivityNet is a large-scale video benchmark of untrimmed videos covering diverse human activities, annotated with temporal activity boundaries and class labels. 
In our construction pipeline, we segment each untrimmed video into temporally localized clips using the provided annotations, and use the corresponding activity class label for each clip as a text signal when building M$^3$KG.

\myparagraph{VALOR}
VALOR is a multimodal dataset of short video clips with synchronized audio and human-authored audio-visual captions, providing closely aligned triplets of vision, audio, and text.
We use the VALOR-32K variant for constructing M$^3$KG.

\subsection{Multimodal RAG Framework} 
In this section, we provide additional details of the multimodal RAG framework used in our experiments. 
Given a multimodal query, we first perform modality-wise retrieval over M$^3$KG. We then apply GRASP, which leverages multimodal grounding models~\cite{liu2024grounding, xu2024towards} and an LLM-based filter implemented with Qwen3-8B ~\cite{yang2025qwen3} using the instruction in Table~\ref{tab:prompt_grasp_filter} to obtain a compact set of query-relevant and answer-supportive triplets, and finally inject this evidence into the MLLM using the graph-augmented generation scheme in Eq. (6) of the main paper.
In practice, Eq.~(6) is realized by using the template summarized in Table~\ref{tab:prompt_rag}.

\begin{table}[t]
\small
\centering
\begin{tabular}{p{0.95\linewidth}}
\toprule
\textbf{LLM-based GRASP filter} \\
\midrule
You are a selector that removes only unnecessary triples for answering the query.

Keep triples that could be helpful to answer the query.

Remove triples that are clearly irrelevant, contradictory to the query, or redundant duplicates.

When uncertain, prefer KEEPING the triple.

Preserve the ORIGINAL ORDER of kept indices (do NOT rerank).

Query: \{QUERY\}

Triplets: \{TRIPLETS\} \\
\bottomrule
\end{tabular}
\caption{Instruction used for the LLM-based filter in GRASP over retrieved triplets.}
\label{tab:prompt_grasp_filter}
\end{table}

\begin{table}[t]
\small
\centering
\begin{tabular}{p{0.95\linewidth}}
\toprule
\textbf{Multimodal RAG Prompt} \\
\midrule
You are a multimodal QA assistant. Prioritize PRIMARY evidence from the input modalities you perceive. \\
Use the retrieved triples BELOW only as optional hints when they are CLEARLY observed or corroborated in the input. \\[4pt]
\texttt{Procedure:} \\
1) Detect whether any triple's context appears in the input (entities, attributes, actions, time/place cues). \\
2) If matched, integrate the FULL triple (head, relation, tail) into the answer, and enrich with head\_desc/tail\_desc. \\
\hspace*{1em}-- Do NOT contradict the primary evidence; if conflict exists, ignore the triple. \\
3) If no triple is confidently matched, answer from the primary evidence only. \\[4pt]
Query : \{QUERY\} \\
Retrieved Triples : \{TRIPLES\_BLOCK\} \\
Triple Format : [i] head=\{h\} | relation=\{r\} | tail=\{t\} || head\_description=\{hd\} | tail\_description=\{td\} \\
Answer : \\
\bottomrule
\end{tabular}
\caption{Graph-augmented generation template used to instantiate Eq.~(6) in our multimodal RAG framework.}
\label{tab:prompt_rag}
\vspace{-1.em}
\end{table}

\begin{table*}[ht]
\small
\centering
\begin{tabular}{p{0.95\linewidth}}
\toprule
\textbf{Reference-Aware Win-rate Prompt} \\
\midrule
You will evaluate two answers to the same question using a Reference Answer. 
Base every judgment solely on alignment to the Reference and the Question; do not reward verbosity or speculative content.

Question:
\{QUESTION\}

Reference Answer (trusted ground truth):
\{REFERENCE\}

Answer 1:
\{ANSWER\_1\}

Answer 2:
\{ANSWER\_2\}

Evaluate on the following criteria and return JSON in the exact schema below.

- Comprehensiveness: Which answer correctly covers more of the Reference's essential points (paraphrase allowed) with fewer mistakes or omissions? Do not reward length; penalize unsupported/contradictory claims.

- Diversity: Which answer offers greater variety in organizing the Reference's facts (e.g., visual vs.\ audio facets) while avoiding new attribute categories not stated or trivially entailed?

- Empowerment: Which answer better enables understanding or action through clear, concise, and reference-aligned guidance (no filler, no meandering)?

- Overall Winner: Choose the answer that is most faithful to the Reference, with stronger correct coverage and clearer, more concise presentation. 
  Break ties by (1) correctness/faithfulness, (2) coverage, (3) concision/clarity. \\

\bottomrule
\end{tabular}
\caption{Reference-aware win-rate comparison template used for LLM-judged preferences.}
\vspace{-1.em}
\label{tab:prompt_winrate}
\end{table*}

\subsubsection{Baselines for Multimodal RAG}
Following prior multimodal RAG work~\cite{park2025vat}, we compare M$^3$KG-RAG against four knowledge-graph baselines coupled with RAG, in addition to the \emph{None} setting where the MLLMs answer without external knowledge.

\myparagraph{Wikidata5M}
Wikidata5M~\cite{wang2021kepler} is a million-scale text-only knowledge graph constructed from Wikidata entities and relations aligned with their textual descriptions from Wikipedia. 

\myparagraph{VTKG}
VTKG~\cite{lee2023vista} is an image-text multimodal knowledge graph that augments a textual KG with visual evidence by attaching images to entities and relational triples, together with short textual descriptions. This design provides entity–relation graphs grounded in visual examples, enabling concept nodes to be linked not only by symbolic relations but also by associated images.

\myparagraph{M$^2$ConceptBase}
M$^2$ConceptBase~\cite{zha2024m2conceptbase} is a concept-centric multimodal knowledge base that represents each concept as a node with multiple aligned visual examples and a detailed textual description. It is explicitly designed to provide fine-grained, cross-modal concept knowledge that can be passed to MLLMs as grounded external evidence, helping mitigate hallucinated or semantically inconsistent predictions.

\myparagraph{VAT-KG}
VAT-KG~\cite{park2025vat} is a knowledge-intensive multimodal knowledge graph that jointly integrates visual, audio, and textual signals into a unified concept-centric graph. Each triplet is linked to multimodal evidence and enriched with concept descriptions, providing an audio-visual KG backbone tailored for retrieval-augmented generation under multimodal queries.

\subsubsection{Evaluation Protocol}

As described in the main paper, our benchmarks consist of open-ended QA with free-form responses, so we adopt an off-the-shelf Model-as-Judge (M.J.) metric~\cite{wang2025audiobench}, where an LLM judge~\cite{grattafiori2024llama} scores each generated answer on a 0-5 scale given the query and reference answer and reports the resulting score on a 0-100 scale.

In addition, we report a pairwise win-rate between M$^3$KG-RAG and each baseline, following RAG evaluation protocols based on LLM preferences~\cite{edge2024local,guo2024lightrag,ren2025videorag}. Unlike prior work that compares two LLM-generated answers, we make the win-rate judge reference-aware by providing the reference answer alongside the two candidates. This helps reduce verbosity bias (overly favoring longer responses) and discourages rewarding merely plausible but unsupported generations, leading to a more faithful and stable preference signal.
The exact evaluation instruction for the reference-aware win-rate judge is provided in Table~\ref{tab:prompt_winrate}. 
The judge compares the two answers according to three criteria—\textit{Comprehensiveness}, \textit{Diversity}, and \textit{Empowerment}—and selects a preferred answer for each criterion. 
Based on these per-criterion preferences, it then decides which answer is preferred overall for each query.

\section{Additional Analysis}
\label{sec:analysis}

\subsection{Hyperparameter Sensitivity Analysis}
\label{sec:hyper}

Our multimodal RAG framework has two scalar hyperparameters: (i) the modality-wise retrieval distance threshold $\tau$ and (ii) the GRASP presence score threshold $\eta$. Both control how much knowledge is injected into the MLLMs and therefore may affect downstream QA performance.
To assess the robustness of our framework to these choices, we conduct a sensitivity study on the VALOR benchmark by varying $\tau$ and $\eta_{av}$ and measuring the resulting M.J. scores.

\myparagraph{Modality-wise distance threshold $\tau$}
For each query, we embed it into the modality-specific representation spaces of M$^3$KG and compute distances to candidate items.
Concretely, an audio-only query is compared against audio items in M$^3$KG within the audio embedding space, and a visual-only query is compared against visual items within the visual embedding space. For an audio-visual query, we concatenate its audio and visual embeddings and match it to audio-visual items in the joint concatenated space.
In each active modality $m$, we compute an L2 distance $d(q_m, x_m)$ between the query representation $q_m$ and an item $x_m$ from M$^3$KG, and use it to perform top-$k$ retrieval.
We then apply the distance threshold~$\tau$ to these $k$ candidates and keep only items with $d(q_m, x_m) \le \tau$. The remaining retrieved items are lifted into the graph via Eq.~(3) in the main paper.

\begin{figure}[t]
    \centering
    \includegraphics[width=0.8\linewidth]{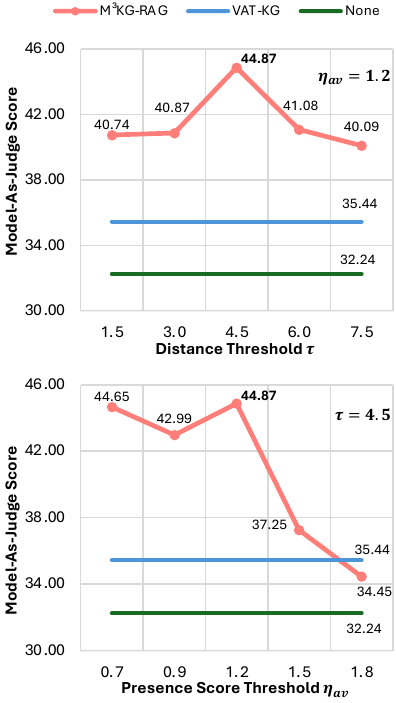}
    \caption{
    \textbf{Sensitivity analysis.}
    M.J.\ score on VALOR versus modality-wise distance threshold $\tau$ (top) and GRASP presence threshold $\eta_{av}$ (bottom).
    }
    \label{fig:supp_graph_ver2}
    \vspace{-2.em}
\end{figure}

To understand how the choice of $\tau$ affects QA performance, we conduct a sensitivity study on the VALOR benchmark by varying $\tau \in \{1.5, 3.0, 4.5, 6.0, 7.5\}$ while fixing $\eta_{av} = 1.2$, and visualize the resulting M.J. scores in the top plot of \cref{fig:supp_graph_ver2}.
As shown in the figure, the M.J. score is maximized at $\tau = 4.5$, while both smaller and larger thresholds yield lower scores. 
Nonetheless, across all tested values of $\tau$, M$^3$KG-RAG consistently outperforms the baselines.
When $\tau$ is set too small, only very close items are retained, so the retrieved subgraphs become overly sparse and may not provide sufficient evidence for the MLLM. 
Conversely, a large $\tau$ allows many more distant items to pass the filter and expands the subgraph, but also introduces multi-hop nodes that are only weakly related to the query, increasing the risk of noisy or distracting knowledge.
These observations are consistent with the intended role of $\tau$ in balancing coverage and noise in modality-wise retrieval.

\vspace{-1.em}
\paragraph{GRASP presence score threshold $\eta$}
GRASP assigns a query-conditioned presence score $s(\mathfrak{t} \mid q)$ to each triplet $\mathfrak{t}$ in the retrieved subgraph using an off-the-shelf multimodal grounding model~\cite{liu2024grounding, xu2024towards}.
We then apply the presence score threshold~$\eta$ and prune triplets whose scores fall below it, keeping only those with $s(\mathfrak{t} \mid q) \ge \eta$.

To examine how $\eta$ affects QA performance, we fix $\tau = 4.5$ and vary $\eta_{av} \in \{0.7, 0.9, 1.2, 1.5, 1.8\}$ on the VALOR benchmark, visualizing the resulting M.J. scores in the bottom plot of \cref{fig:supp_graph_ver2}. The performance is highest at $\eta_{av} = 1.2$, with only minor differences between $\eta_{av} = 0.7$, $0.9$, and $1.2$, but it drops noticeably once $\eta$ increases to $1.5$ and $1.8$. This pattern suggests that moderate grounding-based pruning is beneficial, whereas overly aggressive thresholds remove many triplets that still carry useful evidence, leaving the MLLM with an under-informative subgraph.

\subsection{Additional Ablation Studies}
\label{sec:abl}

\begin{table}[t]
\centering
\small
\setlength{\tabcolsep}{6pt} 
\resizebox{\columnwidth}{!}{
\begin{tabular}{lccc}
\toprule
\textbf{GRASP Component} & \textbf{GPU VRAM (GB)} & \textbf{Avg time / query (s)} & \textbf{M.J.} \\
\midrule
None                        & 23.0   & 4.30    & 40.91 \\
GDino ($\eta_{v}{=}0.8$)      & 23.7 & 5.75  & 41.35 \\
TAG ($\eta_{a}{=}0.4$)        & 23.6 & 4.48  & 41.70 \\
GDino + TAG ($\eta_{av}{=}1.2$) & 24.2 & 6.02  & 42.96 \\
GDino + TAG + LLM Filter           & 39.8 & 7.02 & \textbf{44.87} \\
\bottomrule
\end{tabular}
}
\vspace{-0.5em}
\caption{
\textbf{Ablation on GRASP Components.}
We report GPU VRAM usage, average inference time per query, and M.J. score.
}
\label{tab:exp3_ablation_modules}
\vspace{-2.em}
\end{table}


In Sec.~4.3 of the main paper, we ablate modality-wise retrieval and GRASP as a whole.
We further decompose GRASP into its three submodules in the audio-visual setting: visual grounding with GroundingDINO~\cite{liu2024grounding} (GDino), audio grounding with TAG~\cite{xu2024towards}, and the final LLM-based filtering stage.
\cref{tab:exp3_ablation_modules} reports GPU VRAM, average inference time per query, and M.J. scores on the VALOR benchmark as we progressively enable these components, where GPU VRAM is measured after loading each additional module.

Starting from the configuration without GRASP, adding either GDino or TAG alone yields small but consistent gains over the base M$^3$KG-RAG model (from 40.91 to 41.35 and 41.70 M.J., respectively).
Using both grounding modules together further improves performance to 42.96 M.J., indicating that audio and visual grounding provide complementary benefits.
Importantly, the GPU memory footprint remains almost unchanged when moving from a single grounding module to both (about 23.6–24.2 GB), and the average latency stays within 4.3–6.0 seconds per query.

Finally, enabling the LLM-based filtering stage on top of GDino and TAG achieves the best performance of 44.87 M.J., a gain of nearly 4 points over the configuration without GRASP and about 2 points over using only the grounding modules.
This improvement comes with a moderate increase in resource usage (VRAM from 24.2 GB to 39.8 GB and average time from 6.02 s to 7.02 s per query), while the LLM-based filter helps focus the retrieved subgraph on answer-supporting knowledge.
Overall, these results show that each GRASP submodule contributes positively to performance, and that the full GRASP pipeline offers the best accuracy with a relatively modest overhead compared to its benefits.

\section{Additional Qualitative Results}
\label{sec:qual}

In this section, we present additional qualitative comparisons of M$^{3}$KG-RAG against VAT-KG~\cite{park2025vat} on multimodal QA benchmarks, including Audio QA (AudioCaps-QA~\cite{wang2025audiobench}), Video QA (VCGPT~\cite{maaz2024video}), and Audio-Visual QA (VALOR~\cite{liu2024valor}), using Qwen2.5-Omni~\cite{xu2025qwen2} as the base MLLM. For each benchmark, we show the knowledge retrieved with VAT-KG and with M$^{3}$KG-RAG, the corresponding answers generated from these contexts, and the win-rate judge’s preference and rationale.
Both methods construct MMKGs from raw multimodal corpora and support modality-wise retrieval. However, VAT-KG represents each multimedia item with a single-hop graph and relies on shallow similarity search, often yielding sparse or weakly aligned evidence. In contrast, M$^{3}$KG-RAG exploits multi-hop knowledge and GRASP-based pruning to serve richer, query-relevant context to the MLLM, leading to more faithful and informative responses.

\begin{figure*}[t]
    \centering
    \includegraphics[width=\linewidth]{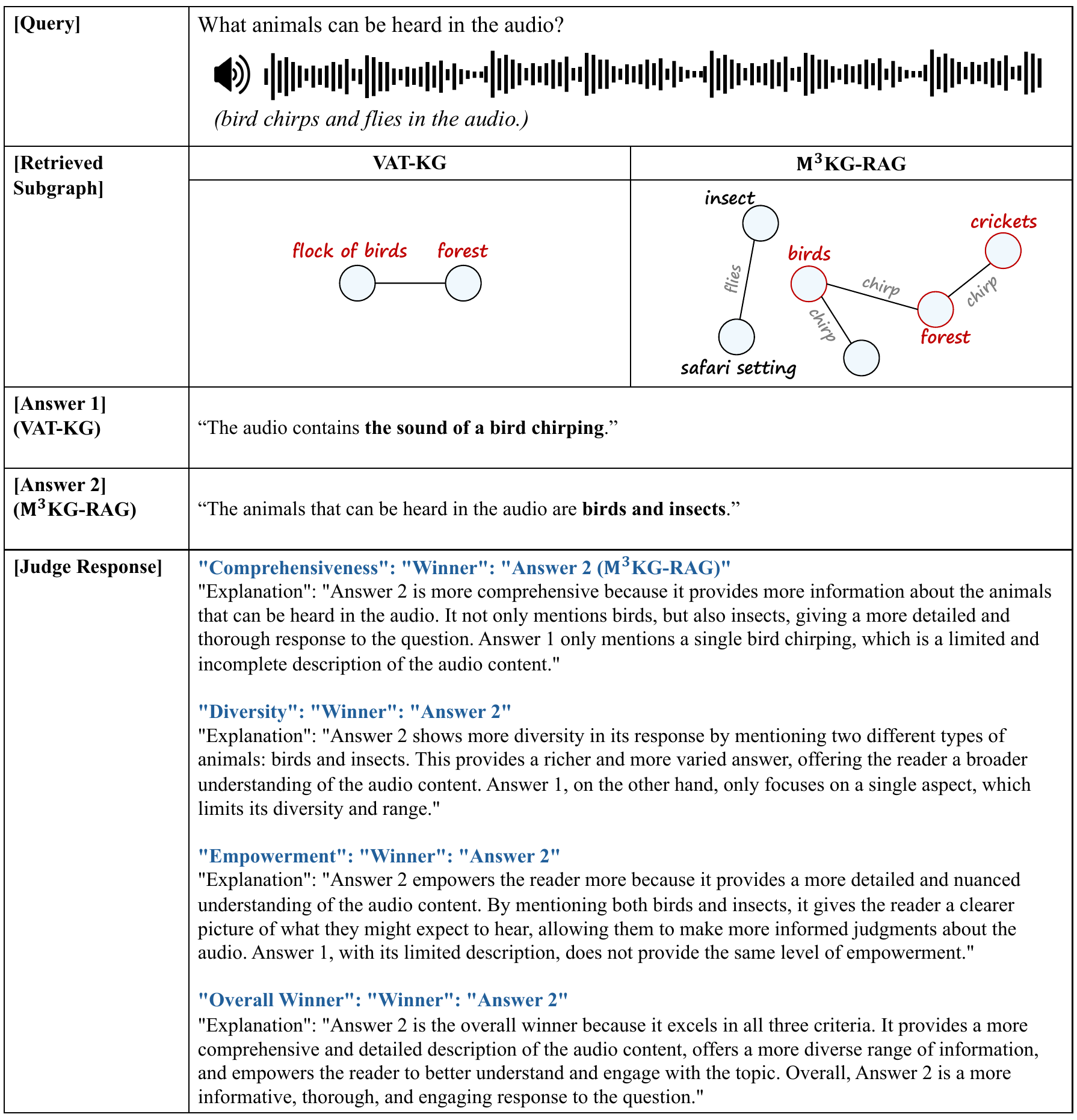}
    \caption{
    \textbf{Qualitative Comparison on Audio QA.}
    Comparing VAT-KG and M$^3$KG-RAG with Qwen2.5-Omni, including retrieved knowledge and win-rate judge preferences.
    }
    \label{fig:supp_aqa}
\end{figure*}

For the Audio QA case in Figure~\ref{fig:supp_aqa}, the query asks what animals can be heard in a clip where birds chirp in a forest environment with background insect sounds. VAT-KG primarily retrieves a single fact about a flock of birds in the forest, leading the model to produce an answer that mentions only birds. In contrast, M$^{3}$KG-RAG retrieves multi-hop knowledge that links both birds and crickets chirping in a forest setting, providing richer cues about co-occurring animal sounds. Conditioned on this context, the model identifies both birds and insects as audible in the scene, which the win-rate judge prefers for covering all relevant animal sources and better matching the reference audio.

\begin{figure*}[t]
    \centering
    \includegraphics[width=\linewidth]{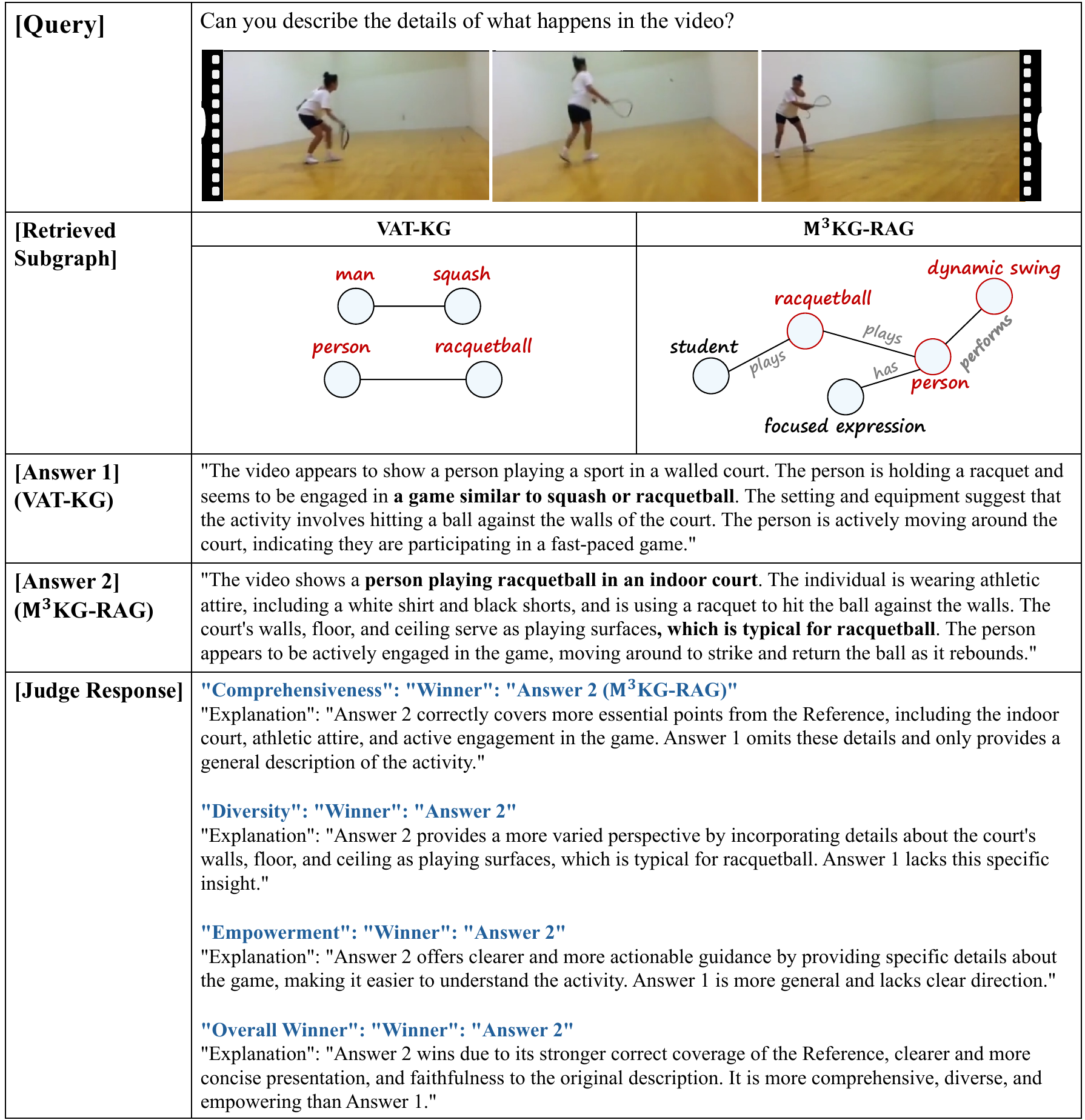}
    \caption{
    \textbf{Qualitative Comparison on Video QA.}
    Comparing VAT-KG and M$^3$KG-RAG with Qwen2.5-Omni, including retrieved knowledge and win-rate judge preferences.
    }
    \label{fig:supp_vcgpt}
\end{figure*}

For the Video QA case in Figure~\ref{fig:supp_vcgpt}, the query video shows a woman playing racquetball on an indoor court, and the model is asked to describe in detail what happens in the scene. VAT-KG performs coarse similarity-based retrieval that includes knowledge about both squash and racquetball, two related but distinct sports, which leads the model to produce a hedged response that refers to a game similar to squash or racquetball without clearly committing to the actual activity or capturing fine-grained details.
In contrast, M$^{3}$KG-RAG, together with GRASP, prunes off-topic neighbors based on fine-grained query relevance and supplies racquetball-focused multi-hop evidence that matches the video.
Guided by this evidence, the model correctly identifies the sport as racquetball and gives a more precise description of the player’s attire, court setting, and actions, which the win-rate judge prefers for its specificity and semantic alignment with the video.

\begin{figure*}[t]
    \centering
    \includegraphics[width=\linewidth]{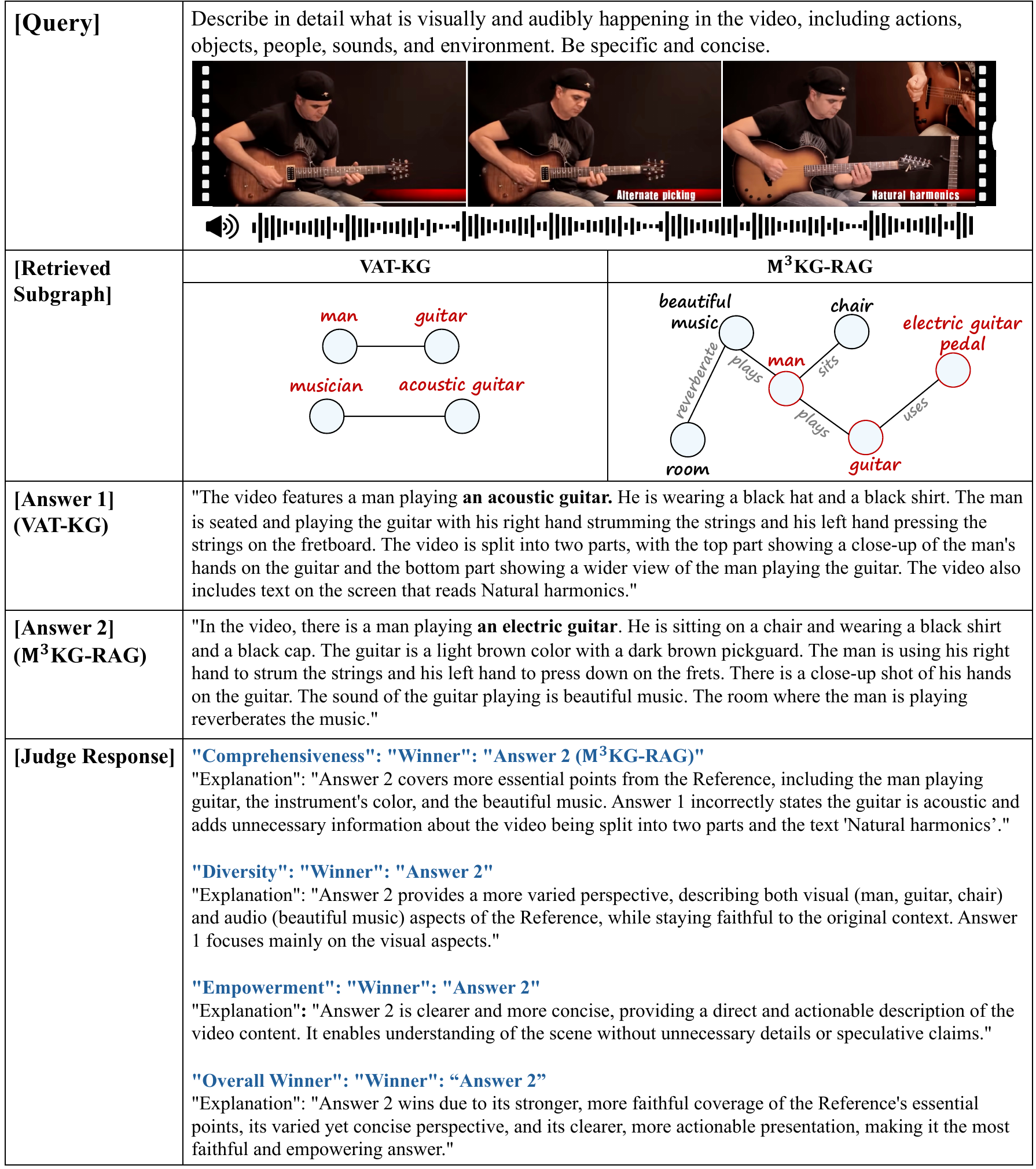}
    \caption{
    \textbf{Qualitative Comparison on Audio-Visual QA.}
    Comparing VAT-KG and M$^3$KG-RAG with Qwen2.5-Omni, including retrieved knowledge and win-rate judge preferences.
    }
    \label{fig:supp_valor}
\end{figure*}

For the Audio-Visual QA case in Figure~\ref{fig:supp_valor}, the multimodal query shows a man playing an electric guitar, and the model is asked to describe the scene. VAT-KG, due to its single-hop MMKG structure, mainly connects the man to a generic guitar and to a musician–acoustic-guitar relation, providing only fragmentary, coarse knowledge. Without fine-grained audio–visual relevance checking, it treats acoustic and electric guitars as semantically interchangeable, which leads the model to describe the scene as an acoustic guitar performance and to miss surrounding contextual details. In contrast, M$^{3}$KG-RAG retrieves a multi-hop neighborhood around the guitar that includes electric-guitar–specific context (such as playing with an effects setup) together with local scene cues (e.g., the man sitting on a chair in an indoor room). With this richer, better-aligned evidence, the model correctly identifies the instrument as an electric guitar and produces a more detailed description of the player’s appearance and environment, which the win-rate judge prefers for both factual correctness and contextual richness.

\section{Limitations}
\label{sec:limit}
Despite the advancements presented in this work, several limitations remain. 
First, the knowledge coverage of M$^{3}$KG is bounded by the raw multimodal corpora used for construction, which may underrepresent long-tail entities, rare relations, or domains not well captured in the source datasets. 
Second, M$^{3}$KG–RAG relies on the multimodal encoders and grounding models used in the retrieval phase; when their training domains do not cover the query distribution, cross-modal miscalibration can surface off-topic neighbors and degrade evidence quality. 
Extending M$^{3}$KG-RAG with better-calibrated encoders, broader and continually updated corpora, and tighter grounding mechanisms is an important direction for future work.

\clearpage


\end{document}